\documentclass[journal]{IEEEtran}
\usepackage{threeparttable}
\usepackage{xcolor,soul,framed} 
\colorlet{shadecolor}{yellow}
\usepackage[pdftex]{graphicx}
\graphicspath{{../pdf/}{../jpeg/}}
\DeclareGraphicsExtensions{.pdf,.jpeg,.png}
\usepackage{subcaption}
\usepackage{adjustbox}
\usepackage{booktabs}
\usepackage{array}
\usepackage{mdwmath}
\usepackage{mdwtab}
\usepackage{eqparbox}
\usepackage{url}
\usepackage{cite}
\usepackage{multirow}
\usepackage{commath}
\usepackage{longtable}
\usepackage{relsize}
\usepackage{amssymb}
\usepackage{amsmath, bm}
\usepackage{bbm}
\usepackage{amsfonts} 
\usepackage[switch]{lineno}
\usepackage{parskip}
\usepackage{rotating}
\usepackage{color}
\usepackage{booktabs,multirow,makecell,rotating,siunitx,adjustbox}
\usepackage{algorithm}
\usepackage{algpseudocode}

\usepackage{subcaption}

\begin{document}
\bstctlcite{IEEEexample:BSTcontrol}
    \title{HypLTSF: A Hyperbolic Geometric View of Multi-Scale Hierarchies for Long-Term Time Series Forecasting}

\author{
\IEEEauthorblockN{
Namwoo Kim\IEEEauthorrefmark{1},
Hyungryul Baik\IEEEauthorrefmark{2},
and Yoonjin Yoon\IEEEauthorrefmark{1}\IEEEauthorrefmark{3}
}
\\[3pt]
\IEEEauthorblockA{
\IEEEauthorrefmark{1}Urban AI Institute, Korea Advanced Institute of Science and Technology (KAIST), Daejeon, Republic of Korea
\\
\IEEEauthorrefmark{2}Department of Mathematical Science, KAIST, Daejeon, Republic of Korea
\\
\IEEEauthorrefmark{3}Department of Civil and Environmental Engineering, KAIST, Daejeon, Republic of Korea
}
\thanks{This work has been submitted to the IEEE for possible publication. Copyright may be transferred without notice, after which this version may no longer be accessible.}
}

\markboth{Preprint}{}


\maketitle

\begin{abstract} 
Multi-scale modeling has become an effective approach for long-term time series forecasting, capturing temporal patterns that range from fine-grained local dynamics to coarse global trends. Representations across these temporal scales are inherently hierarchical, with coarser scales abstracting and aggregating information from finer ones. While existing approaches readily exchange information across these scales, the hierarchy itself is typically left as an emergent byproduct of such interactions rather than captured as a geometric structure in its own right. In this paper, we introduce HypLTSF, a framework that endows the multi-scale hierarchy with a concrete geometric form by embedding scale-wise representations into the Poincaré ball, whose exponentially expanding volume naturally accommodates hierarchical structures. To align this geometry with the temporal hierarchy, HypLTSF imposes two constraints: (1) a radial constraint that orders embeddings by their level of abstraction, and (2) an angular constraint that groups fine-scale patterns sharing a common coarser-scale ancestor. Extensive experiments on long-term time series forecasting benchmarks show that HypLTSF achieves state-of-the-art performance, suggesting that explicitly modeling the multi-scale hierarchy as a geometric structure is effective for forecasting.
\end{abstract}


%


\section{Introduction}

\begin{figure}[t]
    \centering
    \includegraphics[width=\linewidth]{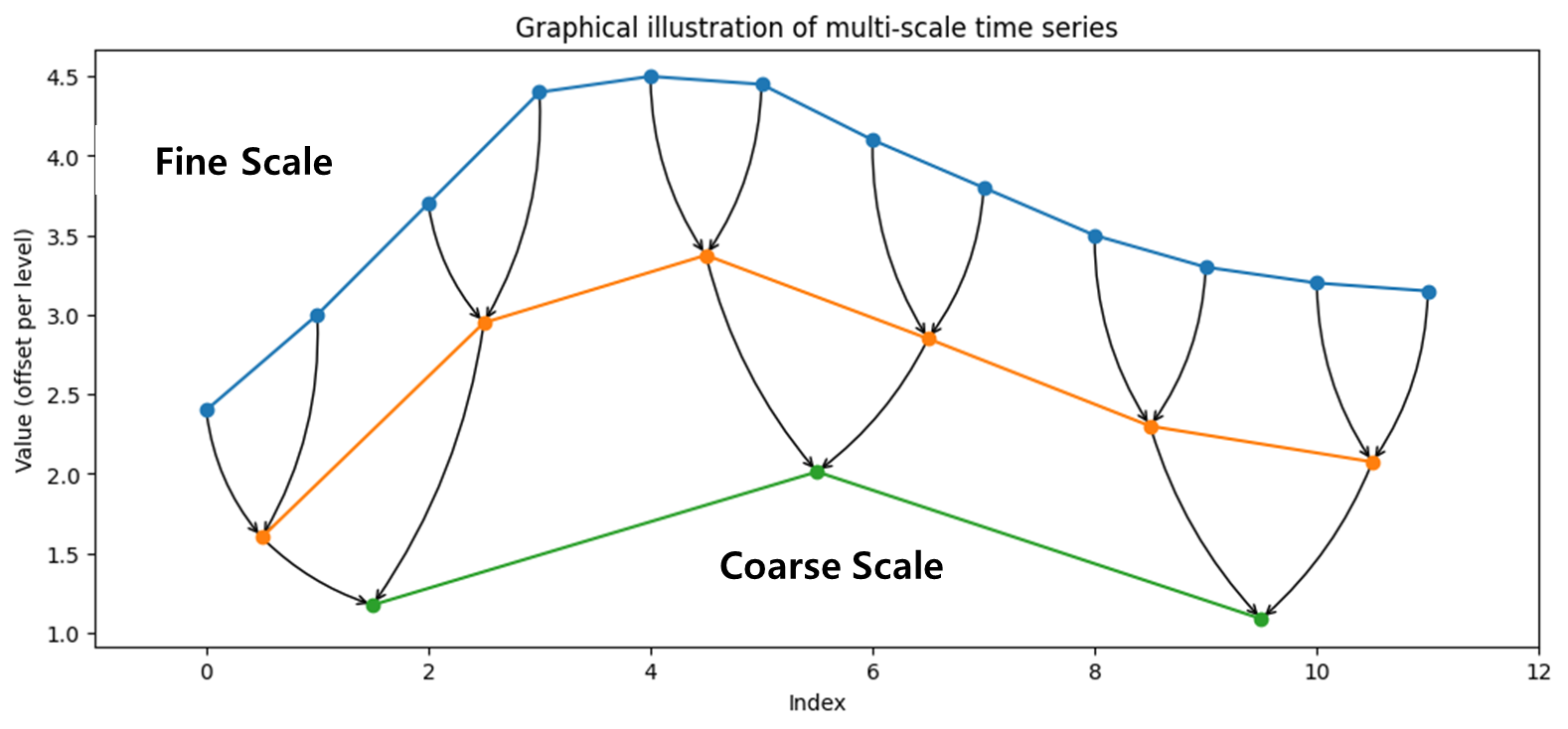}
    \caption{
    Illustration of the hierarchical structure induced by multi-scale representations. Fine-scale observations are progressively aggregated into coarser representations, naturally forming parent--child relationships across scales.
    }
    \label{fig:hierarchy}
\end{figure}

Time series forecasting has attracted significant attention across diverse domains such as economics \cite{huang2024generative, sezer2020financial}, energy systems\cite{chou2018forecasting, wang2023accurate}, and transportation\cite{lippi2013short, yang2021unsupervised}. With the rapid advancement of deep learning, recent research has increasingly focused on learning expressive temporal representations capable of capturing complex and heterogeneous temporal dynamics. One particularly effective paradigm is multi-scale modeling, which represents a time series at multiple temporal resolutions and learns from them jointly~\cite{timemixer,timemixer++,micn,pyraformer,timekan,amd, msdcn}. Operating across resolutions enables models to capture both local and global dynamics. Coarse-scale representations highlight macroscopic behavior, whereas finer-scale representations encode localized variability and short-term dynamics. Beyond providing parallel views of the data, multi-scale representations inherently form a coarse-to-fine hierarchy over time. Existing forecasting methods instantiate this hierarchy by generating a spectrum of representations from raw observations, typically through temporal downsampling \cite{timemixer,timemixer++,timekan,amd} or convolution \cite{micn, msdcn}. As illustrated in Figure~\ref{fig:hierarchy}, repeated application of a downsampling operator or convolution maps fine-scale observations to progressively coarser representations, where each coarse time step aggregates multiple fine-grained observations, naturally inducing parent–child relationships across temporal scales. 

Despite this inherent hierarchy, existing multi-scale methods generally do not explicitly encode hierarchical structure within the geometry of the representation space. They primarily facilitate information exchange across scales via parallel processing, residual aggregation, attention, or mixing operations, while not structurally enforcing the parent–child relationship within the representation space. This limitation raises a natural question: \emph{can we geometrically model the hierarchy underlying multi-scale time series?}. Our framework answers in the affirmative by embedding this hierarchy directly into the geometry of the representation space. We leverage hyperbolic space with a radial ordering constraint to encode abstraction depth. This design places coarse parent representations closer to the origin and fine-grained child representations at larger radii, transforming an implicit hierarchy into an explicit architectural prior. Unlike Euclidean space, where volume grows polynomially with radius, hyperbolic space exhibits exponential volume growth. This property closely mirrors the exponential branching structure of trees. As a result, hyperbolic embeddings can represent hierarchical data with significantly lower distortion than their Euclidean  counterparts~\cite{sarkar2011,nickel2017poincare}. This advantage has led to notable success in domains characterized by latent hierarchies, including knowledge graphs~\cite{chami2020}, natural language~\cite{dhingra2018embedding}, and molecular structures~\cite{liu2019hyperbolic}. However, despite the inherently hierarchical nature of multi-scale temporal representations, hyperbolic geometry has not been widely explored in the context of time series forecasting.

In this paper, we propose \textbf{HypLTSF}, a novel framework that models the hierarchical structure of multi-scale time series in hyperbolic space. We first construct a multi-scale pyramid from the input series via progressive downsampling, decompose each scale into trend and seasonal components, and fuse them into a unified representation. These scale-wise representations are then mapped onto the Poincar\'{e} ball, where we introduce two complementary hierarchy losses to explicitly shape the embedding geometry. The \emph{radial ordering} loss enforces fine-scale embeddings to lie deeper than their coarse-scale counterparts, thereby reflecting the parent–child relationship along the radial axis.
The \emph{angular coherence} loss encourages sibling time steps that are aggregated by the same parent to cluster in similar directions, forming coherent branches in the embedding space. Together, these losses induce a tree-like structure where depth encodes temporal resolution and angular proximity encodes local temporal context. Predictions are generated from each scale and aggregated with learnable weights. Extensive experiments on eight real-world benchmark datasets demonstrate that HypLTSF achieves state-of-the-art forecasting performance. Ablation studies further validate the contributions of hyperbolic embeddings and hierarchy regularization. These results suggest that explicitly modeling the abstraction hierarchy serves as a useful inductive bias that improves multi-scale time series forecasting.

We make the following contributions in this work.
\begin{itemize}
    \item We propose time series forecasting framework that applies hyperbolic geometry to multi-scale long-term time series forecasting, where the hierarchy is induced by progressive temporal downsampling.
    \item We introduce two complementary hierarchy losses in the Poincar\'{e} ball: a radial ordering loss that enforces depth separation between parent and child scales, and an angular coherence loss that clusters sibling time steps together.
    \item Results from comprehensive experiments across eight benchmarks indicate that HypLTSF achieves state-of-the-art performance, which verifies the benefits of explicit hierarchical modeling over implicit approaches. 

\end{itemize}

\section{Related Work}

\subsection{Time Series Forecasting}
Early deep learning approaches for long-term time series forecasting primarily employed RNNs~\cite{deepar,lstnet} and CNNs~\cite{timesnet,scinet} to capture temporal dependencies through recurrent hidden states or convolutional receptive fields. More recently, Transformer- and MLP-based architectures have become the predominant paradigms, demonstrating strong performance.
\paragraph{Transformer-based methods.}
Transformers have been extensively adapted for long-term time series forecasting by redesigning the self-attention mechanism to handle long sequences efficiently.
Informer~\cite{informer} introduced ProbSparse attention to reduce quadratic complexity. Q-Informer~\cite{qinformer} further extends sparse attention with trainable quantum-enhanced query-key scoring. Autoformer~\cite{autoformer} replaced canonical attention with an auto-correlation mechanism coupled with trend--seasonal decomposition.
FEDformer~\cite{fedformer} further improved efficiency by performing attention in the frequency domain. Beyond efficiency, recent work has shifted focus toward the design of input representations and attention targets.  MRformer~\cite{mrformer} captures both long- and short-term temporal patterns by combining global attention with adaptive segment-wise attention. PatchTST~\cite{patchtst} segments time series into subseries-level patches, enabling the model to attend over local semantic units rather than individual time steps. iTransformer~\cite{itransformer} inverts the conventional paradigm by applying self-attention across variates instead of time steps, achieving strong multivariate forecasting performance.
\paragraph{MLP-based methods.}
A parallel line of research has demonstrated that simple MLP architectures can match or even surpass Transformers on standard LTSF benchmarks, challenging the necessity of attention mechanisms.
DLinear~\cite{dlinear} showed that a single linear layer applied after trend--seasonal decomposition achieves surprisingly competitive performance, prompting a fundamental reconsideration of inductive biases in time series models.
TiDE~\cite{tide} extends this direction with an MLP-based encoder-decoder that incorporates covariates while remaining significantly faster than Transformer counterparts.
TSMixer~\cite{tsmixer} alternates time-mixing and feature-mixing MLP layers inspired by MLP-Mixer~\cite{mlpmixer}, and FreTS~\cite{frets} applies MLPs in the frequency domain to exploit the global view and energy compaction properties of spectral representations.
More recently, SparseTSF~\cite{sparsetsf} achieves competitive accuracy with fewer than 1k parameters by decoupling periodicity and trend through cross-period sparse forecasting.

\paragraph{Robust forecasting.} Beyond architectural advances, recent studies have explored improving forecasting robustness against nonstationarity and representation perturbations. JointPGM~\cite{jointpgm} explicitly models time-varying intra-series and inter-series transitional dynamics through a probabilistic graphical framework to address distribution shifts. TopoCL~\cite{topocl} incorporates topological information into contrastive learning to preserve structural properties against augmentation-induced distortions, yielding robust representations applicable to forecasting and other downstream tasks.

\subsection{Hyperbolic Embedding}

Hyperbolic embeddings have been widely studied as an alternative to Euclidean representations for data with inherent hierarchical or tree-like structures. Due to the negative curvature of hyperbolic space, distances grow exponentially with radius, making such spaces particularly well suited for modeling hierarchies, taxonomies, and graphs with power-law degree distributions~\cite{nickel2017poincare}.

In natural language processing, hyperbolic embeddings have been applied to capture semantic hierarchies such as hypernymy relations and lexical taxonomies~\cite{nickel2017poincare,nickel2018learning,ganea2018hyperbolic,dhingra2018embedding}.
More recent work has extended these representations to sentence- and document-level settings~\cite{zhang2021hype,chen2023label}, typically by projecting Euclidean text representations into hyperbolic space or by incorporating hyperbolic distance-based objectives.
These methods have primarily been evaluated on hierarchical text classification and tasks that benefit from explicitly modeling label or semantic hierarchies.

In graph learning, several works have proposed hyperbolic extensions of graph neural networks to model graphs with latent hierarchical structures. HGCN~\cite{HGCN} introduced hyperbolic graph convolutions in the Poincar\'{e} ball, while other works explored Lorentzian geometry~\cite{liu2019hyperbolic} and fully hyperbolic message passing~\cite{dai2021hyperbolic}. HTGN~\cite{htgn} embeds discrete-time temporal graphs in hyperbolic space to jointly capture temporal evolution and latent hierarchy.

In this work, we target continuous multivariate time series, where no relational structure is given a priori and the hierarchy is instead induced by a temporal downsampling operator. To the best of our knowledge, hyperbolic geometry has not been widely studied in the context of long-term time series forecasting.

\subsection{Multi-Scale Modeling}
Multi-scale modeling has emerged as an effective paradigm for time series forecasting, motivated by the observation that temporal dependencies manifest across multiple time resolutions. Real-world time series typically exhibit a mixture of long-term trends, seasonal patterns, and short-term fluctuations, which are difficult to capture within a single temporal scale.

Existing multi-scale approaches derive multi-resolution (often coarse-to-fine) representations from raw data through various mechanisms.
Downsampling-based methods such as TimeMixer~\cite{timemixer}, TimeMixer++~\cite{timemixer++}, and N-HiTS~\cite{nhits} construct explicit temporal pyramids, where coarser scales summarize or parameterize finer observations.
Pyraformer~\cite{pyraformer} builds multi-resolution representations by constructing coarse-scale nodes using convolutions and enables information exchange across scales through pyramidal sparse attention. Ada-MSHyper~\cite{adamshyper} models multi-scale temporal dependencies via adaptive hypergraph learning and scale-aware hypergraph interactions. Convolution-based methods like MICN~\cite{micn} employ varying kernel sizes to capture patterns at different temporal granularities, producing coarser representations through larger receptive fields.
Similarly, ResMMoT-Informer~\cite{resmmot} employs heterogeneous TCN experts with different receptive fields and sparse expert selection to adaptively capture temporal patterns across multiple scales.

These approaches collectively span a fine-to-coarse spectrum of temporal representations. However, while these methods exchange information across scales, they do not impose structural constraint that enforces a hierarchical ordering among scales. On the other hand, proposed framework models the hierarchical relationship directly as a geometric constraint in hyperbolic space. Specifically, we embed multi-scale representations in hyperbolic space, where a radial ordering constraint directly encodes abstraction depth, ensuring that fine-grained (child) representations lie deeper than their coarse (parent) counterparts.

\section{Preliminaries}
\subsection{Poincar\'e ball}
We briefly review the Poincar\'e ball model of hyperbolic space.
Let $c>0$ denote the curvature parameter, corresponding to constant sectional curvature $-c$.
The $d$-dimensional Poincar\'e ball is
\begin{equation}
\mathbb{D}_c^d \;=\; \bigl\{ \mathbf{x}\in\mathbb{R}^d : c\|\mathbf{x}\|^2 < 1 \bigr\},
\end{equation}
equipped with the conformal Riemannian metric
\begin{equation}
g_{\mathbf{x}}^c = \lambda_{\mathbf{x}}^2 \, g^E,
\qquad
\lambda_{\mathbf{x}} = \frac{2}{1-c\|\mathbf{x}\|^2},
\end{equation}
where $g^E$ is the Euclidean metric.
The geodesic distance between $\mathbf{x},\mathbf{y}\in\mathbb{D}_c^d$ is
\begin{equation}\label{eq:hyp-dist}
d_c(\mathbf{x},\mathbf{y})
= \frac{2}{\sqrt{c}}\,\operatorname{artanh}\!\bigl(\sqrt{c}\,\|-\mathbf{x}\oplus_c \mathbf{y}\|\bigr),
\end{equation}
where $\oplus_c$ is M\"obius addition (defined below).
A key property is that the distance from the origin reduces to
\begin{equation}\label{eq:hyp-depth}
d_c(\mathbf{0},\mathbf{x})
= \frac{2}{\sqrt{c}}\,\operatorname{artanh}(\sqrt{c}\|\mathbf{x}\|),
\end{equation}
which serves as a natural measure of \emph{hierarchical depth}.

\subsection{Operations in hyperbolic space}

\paragraph{M\"obius addition and scalar multiplication.}
For $\mathbf{x},\mathbf{y}\in\mathbb{D}_c^d$, M\"obius addition is
\begin{equation}
\mathbf{x}\oplus_c \mathbf{y}
= \frac{(1+2c\langle\mathbf{x},\mathbf{y}\rangle+c\|\mathbf{y}\|^2)\mathbf{x}+(1-c\|\mathbf{x}\|^2)\mathbf{y}}{1+2c\langle\mathbf{x},\mathbf{y}\rangle+c^2\|\mathbf{x}\|^2\|\mathbf{y}\|^2}.
\end{equation}
M\"obius scalar multiplication for $r\in\mathbb{R}$ and $\mathbf{x}\neq\mathbf{0}$ is
\begin{equation}
r\otimes_c \mathbf{x}
= \frac{1}{\sqrt{c}}\tanh\!\bigl(r\,\operatorname{artanh}(\sqrt{c}\|\mathbf{x}\|)\bigr)\frac{\mathbf{x}}{\|\mathbf{x}\|},
\qquad r\otimes_c \mathbf{0}=\mathbf{0}.
\end{equation}

\paragraph{Exponential and logarithmic maps.}
The exponential map at $\mathbf{x}\in\mathbb{D}_c^d$ sends a tangent vector $\mathbf{v}\in T_{\mathbf{x}}\mathbb{D}_c^d$ to the manifold:
\begin{equation}
\label{eq:expmap}
\begin{aligned}
\exp_{\mathbf{x}}^c(\mathbf{v})
&= \mathbf{x}\oplus_c\!\left(
\frac{1}{\sqrt{c}}
\tanh\!\left(
\frac{\sqrt{c}\,\lambda_{\mathbf{x}}\|\mathbf{v}\|}{2}
\right)
\frac{\mathbf{v}}{\|\mathbf{v}\|}
\right),\\
\exp_{\mathbf{x}}^c(\mathbf{0})
&= \mathbf{x}.
\end{aligned}
\end{equation}
The logarithmic map $\log_{\mathbf{x}}^c:\mathbb{D}_c^d\to T_{\mathbf{x}}\mathbb{D}_c^d$ is
\begin{equation}\label{eq:logmap}
\log_{\mathbf{x}}^c(\mathbf{y})
= \frac{2}{\sqrt{c}\,\lambda_{\mathbf{x}}}\operatorname{artanh}\!\bigl(\sqrt{c}\,\|-\mathbf{x}\oplus_c\mathbf{y}\|\bigr)\frac{-\mathbf{x}\oplus_c\mathbf{y}}{\|-\mathbf{x}\oplus_c\mathbf{y}\|}.
\end{equation}

\paragraph{Einstein midpoint.}
The Einstein midpoint provides a hyperbolic analogue of the Euclidean centroid. Given points $\{\mathbf{x}_i\}_{i=1}^{n}\subset\mathbb{D}_c^d$, it is defined as
\begin{equation}\label{eq:einstein}
\bar{\mathbf{x}}=\frac{\sum_{i=1}^{n}\gamma_i\,\mathbf{x}_i}{\sum_{i=1}^{n}\gamma_i},\qquad \gamma_i=\frac{1}{\sqrt{1-c\|\mathbf{x}_i\|^2}},
\end{equation}
where $\gamma_i$ is the Lorentz factor, which assigns higher weight to points closer to the boundary to compensate for the rapid growth of hyperbolic distances in that region.

\subsection{Problem Statement}
Multivariate Time Series Forecasting is the task of predicting future values of multiple variables based on their historical observations. Let $\mathbf{X} = \{\mathbf{x}_1, \mathbf{x}_2, \ldots, \mathbf{x}_T\}$ denote a multivariate time series, where $\mathbf{x}_t \in \mathbb{R}^C$ represents the observations of $C$ variables at time step $t$. Given a look-back window of length $L$, the input at time $t$ is defined as $\mathbf{X}_t = [\mathbf{x}_{t-L+1}, \ldots, \mathbf{x}_t] \in \mathbb{R}^{C \times L}$.

The objective is to learn a forecasting function $f: \mathbb{R}^{C \times L} \rightarrow \mathbb{R}^{C \times H}$ that maps the historical observations to future predictions $\hat{\mathbf{Y}}_t = [\hat{\mathbf{x}}_{t+1}, \ldots, \hat{\mathbf{x}}_{t+H}] \in \mathbb{R}^{C \times H}$, where $H$ denotes the prediction horizon. The function $f$ is optimized to minimize the discrepancy between the predicted values $\hat{\mathbf{Y}}_t$ and the ground truth $\mathbf{Y}_t = [\mathbf{x}_{t+1}, \ldots, \mathbf{x}_{t+H}]$.

\section{Methods}

\begin{figure*}[t]
    \centering
    \includegraphics[width=\textwidth]{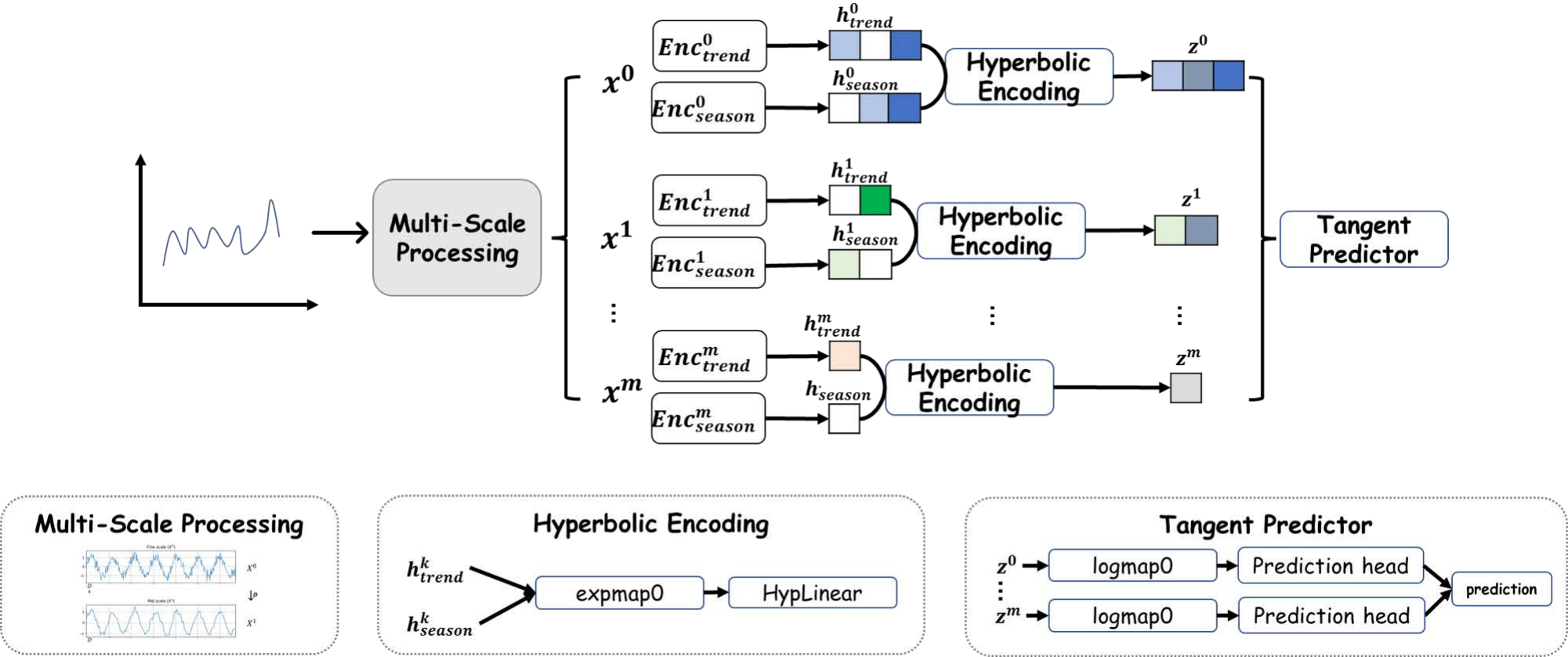}
    \caption{Overview of the HypLTSF. We build a multi-scale pyramid via progressive downsampling, decompose each scale into trend and seasonality, and fuse them to obtain scale-wise features. The scale-wise features are then mapped to the Poincaré ball via hyperbolic encoding, where two complementary hierarchy losses shape the embedding geometry. Finally, predictions are generated in the tangent space.
    }
    \label{fig:overview}
\end{figure*}

We propose HypLTSF, a multi-scale forecasting framework that impose structural constraint that explicitly enforces a hierarchical ordering among scales. Our central hypothesis is that the multi-scale structure of time series---where fine-grained fluctuations aggregate into coarser trends---forms a natural hierarchy that can be captured by the radial and angular geometry of hyperbolic space. For better understanding, overview of the HypLTSF can be found in Figure~\ref{fig:overview}. 

\subsection{Multi-scale Decomposition}

\paragraph{Scale construction.}
Given input $\mathbf{x}\in\mathbb{R}^{C\times L}$, we construct $S{+}1$ scales by progressive average pooling:
\begin{equation}
\mathbf{x}^{(1)}=\mathbf{x},\qquad
\mathbf{x}^{(s+1)}=\operatorname{AvgPool}_w\!\bigl(\mathbf{x}^{(s)}\bigr),\quad s=1,\dots,S,
\end{equation}
where $w$ is the downsampling window.
This produces a pyramid $\{\mathbf{x}^{(s)}\}_{s=1}^{S+1}$ with $\mathbf{x}^{(s)}\in\mathbb{R}^{C \times L_s}$ and $L_s=L/w^{s-1}$.

\paragraph{Normalization.}
Each scale is independently normalized with learnable affine parameters to handle distribution shift across resolutions \cite{revin}:
\begin{equation}
\tilde{\mathbf{x}}^{(s)}=\gamma^{(s)}\odot\frac{\mathbf{x}^{(s)}-\mu^{(s)}}{\sigma^{(s)}}+\beta^{(s)},
\end{equation}
where $\mu^{(s)},\sigma^{(s)}$ are per-instance statistics along the time dimension. Here, $\gamma^{(s)}$ and $\beta^{(s)}$ are learnable scale and shift parameters, respectively, which allow the model to adaptively rescale and shift the normalized representations at each scale.
This normalization is inverted after prediction to recover the original scale.

\paragraph{Trend--seasonal decomposition.}
Following classical time series analysis~\cite{cleveland1990stl}, the channel-independent representation at each scale is decomposed into trend and seasonal components:
\begin{equation}
\mathbf{x}^{(s)}_{\mathrm{trend}}=\operatorname{MovingAvg}_k\!\bigl(\tilde{\mathbf{x}}^{(s)}\bigr),\qquad
\mathbf{x}^{(s)}_{\mathrm{season}}=\tilde{\mathbf{x}}^{(s)}-\mathbf{x}^{(s)}_{\mathrm{trend}}.
\end{equation}

\subsection{Euclidean Encoding}
Each trend and seasonal component is processed by a dedicated encoder $\operatorname{Enc}^{(s)}$ that captures both temporal and cross-feature dependencies in two successive stages.

The first stage operates along the time axis with a scale-specific two-layer MLP.
Because each scale $s$ has a different temporal length $L_s = L / w^{s-1}$, we use separate weight matrices per scale to model resolution-appropriate temporal patterns:
\begin{equation}
\tilde{\mathbf{x}}^{(s)} = W_2^{(s)}\,\operatorname{GELU}\!\bigl(W_1^{(s)}\,{\mathbf{x}^{(s)}}^\top\bigr)^\top,
\end{equation}
where $W_1^{(s)},W_2^{(s)}\in\mathbb{R}^{L_s\times L_s}$.

The second stage operates along the model dimension with a two-layer MLP that is shared across all scales, enabling cross-feature interaction and ensuring a consistent representation space regardless of temporal resolution:
\begin{equation}
\mathbf{h}^{(s)} = W_4\,\operatorname{GELU}\!\bigl(W_3\,\tilde{\mathbf{x}}^{(s)}\bigr),
\end{equation}
where $W_3\in\mathbb{R}^{d_\mathrm{ff}\times d_\mathrm{model}}$ and $W_4\in\mathbb{R}^{d_\mathrm{model}\times d_\mathrm{ff}}$.

Finally, separate encoders with independent temporal weights but shared feature weights are applied to each season and trend component:
\begin{equation}
\mathbf{h}^{(s)}_{\mathrm{season}}=\operatorname{Enc}^{(s)}_{\mathrm{season}}\!\bigl(\mathbf{x}^{(s)}_{\mathrm{season}}\bigr),\qquad
\mathbf{h}^{(s)}_{\mathrm{trend}}=\operatorname{Enc}^{(s)}_{\mathrm{trend}}\!\bigl(\mathbf{x}^{(s)}_{\mathrm{trend}}\bigr).
\end{equation}

\subsection{Hyperbolic Encoding}

The encoded representations live in Euclidean space and carry no notion of hierarchy.
We map them into the Poincar\'e ball $\mathbb{D}_c^{d_h}$, whose radial geometry naturally encodes hierarchical depth (Eq.~\ref{eq:hyp-depth}).

\paragraph{Component fusion.}
Trend and seasonal encodings are fused in Euclidean space with learnable per-scale weights before mapping onto the ball:
\begin{equation}
\mathbf{f}^{(s)}=\operatorname{LayerNorm}\!\bigl(w_1^{(s)}\,\mathbf{h}^{(s)}_{\mathrm{season}}+w_2^{(s)}\,\mathbf{h}^{(s)}_{\mathrm{trend}}\bigr),
\end{equation}
where $w_1^{(s)},w_2^{(s)}$ are learnable scalars that allow the model to adapt the relative importance of each component at each scale.

\paragraph{Hyperbolic projection.}
The fused representation is first mapped onto $\mathbb{D}_c^{d_h}$ via the exponential map at the origin (Eq.~\ref{eq:expmap} with $\mathbf{x}=\mathbf{0}$), then transformed by a scale-specific hyperbolic linear layer:
\begin{equation}
\mathbf{z}^{(s)}=\operatorname{HypLinear}^{(s)}\!\bigl(\exp_{\mathbf{0}}^c(\mathbf{f}^{(s)})\bigr).
\end{equation}
The hyperbolic linear layer maps $d_\mathrm{model}\to d_h$ via M\"obius matrix-vector multiplication and bias addition:
\begin{equation}
\operatorname{HypLinear}(\mathbf{x})=M\otimes_c\mathbf{x}\oplus_c\exp_{\mathbf{0}}^c(\mathbf{b}),
\label{eq:hyplinear}
\end{equation}
where $M\in\mathbb{R}^{d_h\times d_\mathrm{model}}$ is a learnable weight matrix and $\mathbf{b}\in\mathbb{R}^{d_h}$ is a learnable bias mapped onto the ball via the exponential map before M\"obius addition.
The M\"obius matrix-vector product is defined as
\begin{equation}
M\otimes_c\mathbf{x}=\frac{1}{\sqrt{c}}\tanh\!\Bigl(\frac{\|M\mathbf{x}\|}{\|\mathbf{x}\|}\operatorname{artanh}(\sqrt{c}\|\mathbf{x}\|)\Bigr)\frac{M\mathbf{x}}{\|M\mathbf{x}\|}.
\end{equation}
In practice, a projection that clips the norm to $(1{-}\epsilon)/\sqrt{c}$ is applied after each M\"obius operation to ensure numerical stability. The result is a set of hyperbolic embeddings $\{\mathbf{z}^{(s)}\}_{s=1}^{S+1}$. Then, our hierarchy losses (Section~\ref{sec:hier}) shape these embeddings to impose a hierarchical ordering across scales, with finer scales embedded deeper than coarser scales.

\subsection{Hierarchy Losses}\label{sec:hier}

The downsampling procedure creates natural parent--child relationships across scales: each time step at scale $s{+}1$ (parent) corresponds to $w$ consecutive time steps at scale $s$ (children):
\begin{equation}
\operatorname{children}(t,s{+}1)=\bigl\{\mathbf{z}^{(s)}_{wt},\;\mathbf{z}^{(s)}_{wt+1},\;\ldots,\;\mathbf{z}^{(s)}_{wt+w-1}\bigr\}.
\end{equation}
Each scale embedding has shape $C\times L_s\times d_h$.
To compute the hierarchy losses, we first aggregate across channels via the Einstein midpoint (Eq.~\ref{eq:einstein}), yielding a channel-aggregated representation $\bar{\mathbf{z}}^{(s)}\in\mathbb{R}^{L_s\times d_h}$ on which all subsequent loss computations are performed.
We enforce the hierarchy along two complementary geometric axes.

\paragraph{Radial ordering.}
In the Poincar\'e ball, hierarchical depth is encoded by distance from the origin (Eq.~\ref{eq:hyp-depth}).
We require that every child individually lies deeper than its parent by a learnable margin $m$:
\begin{equation}\label{eq:radial}
\begin{aligned}
\mathcal{L}_{\mathrm{r}}
&=\frac{1}{S}\sum_{s=1}^{S}\frac{1}{L_{s+1}\cdot w}
\sum_{t}\sum_{j=0}^{w-1}
\Bigl[\max\!\bigl(0,\;
d_c(\mathbf{0},\bar{\mathbf{z}}^{(s+1)}_t) \\
&\hspace{7.5em}
-d_c(\mathbf{0},\bar{\mathbf{z}}^{(s)}_{wt+j})+m
\bigr)\Bigr]^2 .
\end{aligned}
\end{equation}
The squared hinge provides a strong gradient signal on large violations while contributing zero loss when the constraint is already satisfied.
The learnable margin allows the model to discover the appropriate radial separation for each dataset.

\paragraph{Angular coherence.}
Radial ordering establishes hierarchy levels but leaves the angular arrangement unconstrained---points at the correct depth could scatter in arbitrary directions, which does not constitute a tree.
To induce branch structure, we measure angular coherence in the \emph{tangent space at the parent}, which provides a geometrically principled local coordinate system.

For parent $t$ at scale $s{+}1$, we compute the tangent vector from the parent to each child via the logarithmic map (Eq.~\ref{eq:logmap}):
\begin{equation}
\mathbf{v}_{t,j}=\log_{\bar{\mathbf{z}}^{(s+1)}_t}^c\!\bigl(\bar{\mathbf{z}}^{(s)}_{wt+j}\bigr),\qquad j=0,\ldots,w{-}1.
\end{equation}
The \emph{branch direction} is defined as the normalised mean of these tangent vectors:
\begin{equation}
\hat{\mathbf{b}}_t=\frac{\sum_{j=0}^{w-1}\mathbf{v}_{t,j}}{\bigl\|\sum_{j=0}^{w-1}\mathbf{v}_{t,j}\bigr\|}.
\end{equation}
The angular coherence loss pulls each child's tangent direction toward the branch direction:
\begin{equation}\label{eq:angular}
\mathcal{L}_{\mathrm{a}}=\frac{1}{S}\sum_{s=1}^{S}\;\frac{1}{L_{s+1}\cdot w}\sum_{t}\sum_{j=0}^{w-1}\Bigl(1-\bigl\langle\hat{\mathbf{v}}_{t,j},\;\hat{\mathbf{b}}_t\bigr\rangle\Bigr),
\end{equation}
where $\hat{\mathbf{v}}=\mathbf{v}/\|\mathbf{v}\|$.
By operating in the tangent space at the parent rather than using global directions from the origin, this formulation respects the local geometry of the Poincar\'e ball and produces more meaningful angular comparison.

\begin{figure}[t]
  \centering
  \subfloat[Radial Ordering Loss]{\includegraphics[width=0.45\columnwidth]{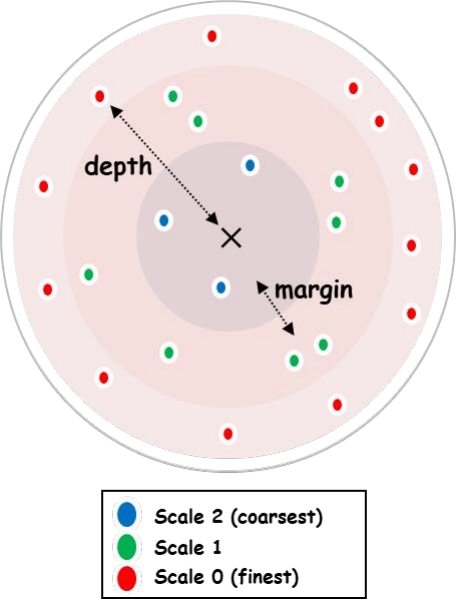}}
  \hfill
  \subfloat[Angular Coherence Loss]{\includegraphics[width=0.45\columnwidth]{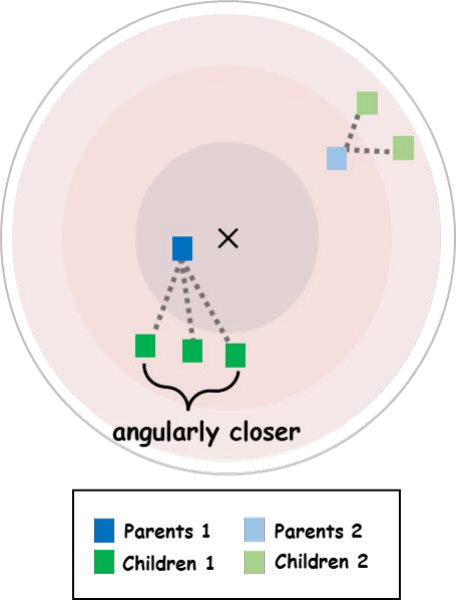}}
  \caption{Illustration of the two hierarchy-preserving losses in the Poincar\'e disk. (a) The radial ordering loss enforces that coarse-scale (parent) embeddings lie relatively closer to the origin than fine-scale (child) embeddings, with their radii separated by a learnable margin $m$. (b)~The angular coherence loss encourages children of the same parent to be angularly close in the tangent space at the parent, forming coherent branches.}
  \label{fig:hierarchy_loss}
  \end{figure}

\paragraph{Joint effect.}
The two losses shape complementary axes of the Poincar\'e ball:
\begin{itemize}
\item $\mathcal{L}_{\mathrm{r}}$ organizes the \emph{radial} axis—children are positioned deeper than their parents, closer to the boundary.
\item $\mathcal{L}_{\mathrm{a}}$ organizes the \emph{angular} axis---siblings cluster in the tangent space at their parent, forming coherent subtrees.
\end{itemize}
Together, they induce a tree-like embedding where hierarchical depth corresponds to temporal resolution and angular proximity corresponds to local temporal context. For better understanding, graphical illustration of hierarchy losses is shown in Figure \ref{fig:hierarchy_loss}.

\subsection{Prediction}

Predictions are generated in the tangent space at the origin, which is locally Euclidean and amenable to standard linear operations.

\paragraph{Tangent-space predictor.}
For each scale, we map hyperbolic embeddings back via the logarithmic map (Eq.~\ref{eq:logmap} with $\mathbf{x}=\mathbf{0}$) and apply scale-specific projection:
\begin{equation}
\hat{\mathbf{y}}^{(s)}=W_\mathrm{dim}\Bigl(W^{(s)}_\mathrm{time}\operatorname{GELU}\bigl(\log{\mathbf{0}}^c(\mathbf{z}^{(s)})\bigr)\Bigr)^{\top},
\end{equation}
where $W^{(s)}_\mathrm{time}\in\mathbb{R}^{H\times L_s}$ projects along the temporal dimension and $W_\mathrm{dim}\in\mathbb{R}^{1\times d_h}$ reduces to univariate predictions per channel.

\paragraph{Multi-scale aggregation.}
The final forecast is a learnable weighted combination of all scales:
\begin{equation}
\hat{\mathbf{y}}=\sum_{s=1}^{S+1}w^{(s)}\,\hat{\mathbf{y}}^{(s)},\qquad w^{(s)}=\frac{\exp(\delta^{(s)})}{\sum_{s'}\exp(\delta^{(s')})},
\end{equation}
where $\{\delta^{(s)}\}$ are learnable parameters.

\subsection{Training Objective}

The complete objective combines the prediction loss with the hierarchy constraints:
\begin{equation}
\mathcal{L}=\mathcal{L}_{\mathrm{pred}}+\lambda_r\,\mathcal{L}_{\mathrm{r}}+\lambda_a\,\mathcal{L}_{\mathrm{a}},
\end{equation}
where $\lambda_r$ and $\lambda_a$ control the strength of each regularizer. For $\mathcal{L}_{\mathrm{pred}}$, the L1 loss is employed. The hierarchy losses serve as inductive biases: rather than discovering hierarchical structure from the prediction signal alone, the model is guided toward representations where temporal resolution maps to hyperbolic depth and local context maps to angular position.

\section{Experiments}

\subsection{Experimental Setup}

We evaluate HypLTSF on eight widely used forecasting benchmarks, including ETT, Weather, Electricity, Traffic, and Solar. The baselines cover a broad range of forecasting methods, including Amplifier~\cite{amplifier}, iTransformer~\cite{itransformer}, PatchTST~\cite{patchtst}, Crossformer~\cite{crossformer}, TimesNet~\cite{timesnet}, DLinear~\cite{dlinear}, FEDformer~\cite{fedformer}, and Non-stationary Transformer (Stationary)~\cite{stationary}. Recent multi-scale approaches, including RAFT~\cite{raft}, TimeKAN~\cite{timekan}, and TimeMixer~\cite{timemixer}, are also included for comparison. This diverse selection enables a thorough comparison across fundamentally different modeling strategies. 

Following ~\cite{tfb, duet, srsnet}, we do not apply the drop-last strategy to ensure fair comparison. All methods are compared at four prediction horizons, namely $96$, $192$, $336$, and $720$, with MSE and MAE as evaluation metrics. To account for the sensitivity of different models to the input context length, we consider multiple look-back window sizes $L \in \{96, 336, 512\}$ for all datasets and report each method’s best performance. This protocol ensures both consistent evaluation and fair model comparison across varying temporal dependencies. All experiments are conducted using PyTorch with Python~3.8 and the Adam optimizer on a single NVIDIA RTX~3090 GPU. Each experiment is repeated over three independent runs. 


\subsection{Main Results}

Table~\ref{tab:main_results} summarizes the forecasting performance, with the top and runner-up results in each row marked in bold red and underlined blue, respectively. The \emph{Avg} row reports the per-dataset mean across the four prediction horizons, providing a compact summary of overall performance on each benchmark. From these results, we make the following observations.

First, HypLTSF achieves the strongest overall performance across the eight benchmarks, with a particularly pronounced advantage in MAE. At the dataset-average level, it attains the lowest MAE on all eight datasets and the lowest MSE on six. This advantage is even more evident across forecasting horizons: among the 32 settings, HypLTSF ranks first in 30 cases for MAE and 20 for MSE. The stronger gains in MAE, together with consistently competitive MSE results, indicate that the proposed multi-scale hyperbolic hierarchy provides robust improvements across different error criteria. Notably, the improvements hold across both the ETT benchmarks and larger-scale datasets, including Weather, Electricity, Solar, and Traffic, demonstrating consistent benefits across diverse forecasting regimes.

Second, compared with recent baselines, HypLTSF remains competitive with TimeKAN and RAFT on the ETT benchmarks, where performance differences among the top methods are relatively small. While TimeKAN achieves slightly lower average MSE on ETTh2 (0.350 vs. 0.352) and ETTm1 (0.344 vs. 0.347), HypLTSF consistently yields the lowest average MAE across all four ETT datasets. Its advantage becomes more pronounced on the larger datasets. On Solar at $H{=}720$, HypLTSF reduces MSE by 4.8\% (0.209$\to$0.199) and MAE by 16.4\% (0.269$\to$0.225) relative to TimeKAN. Similarly, on Traffic, it achieves the best average MAE, improving upon the strongest competing baseline by 6.5\% (0.275$\to$0.257). These results indicate that the benefits of the proposed hyperbolic hierarchy remain consistent across datasets and become particularly pronounced in larger, high-dimensional forecasting settings.

Third, against TimeMixer, a representative multi-scale forecasting model that shares the same MLP-based backbone, HypLTSF exhibits consistent dataset-average improvements, with notable gains on Solar, where MAE is reduced by 14.7\% (0.252$\to$0.215), and on Traffic, where MAE is reduced by 7.9\% (0.279$\to$0.257). Given their similar architectures, these results suggest that explicitly modeling the multi-scale hierarchy in hyperbolic space provides an effective geometric prior for organizing representations across temporal scales.

\begin{table*}[h]
\caption{Multivariate forecasting results with forecasting horizons $H \in \{96, 192, 336, 720\}$ for the datasets.
We consider multiple look-back window sizes $L \in \{96, 336, 512\}$ for all datasets and report each method’s best performance. The Avg row reports the average MSE/MAE across the four horizons for each dataset.}
\label{tab:main_results}
\resizebox{\textwidth}{!}{%
\begin{tabular}{cc|cc|cc|cc|cc|cc|cc|cc|cc|cc|cc|cc|cc}
\toprule
\multicolumn{2}{c|}{\multirow{2}{*}{Models}} & \multicolumn{2}{c}{HypLTSF} &\multicolumn{2}{c}{TimeKAN} &\multicolumn{2}{c}{Amplifier} &\multicolumn{2}{c}{RAFT} &\multicolumn{2}{c}{iTransformer} &\multicolumn{2}{c}{TimeMixer} &\multicolumn{2}{c}{PatchTST} &\multicolumn{2}{c}{Crossformer} &\multicolumn{2}{c}{TimesNet} &\multicolumn{2}{c}{DLinear} &\multicolumn{2}{c}{Stationary} &\multicolumn{2}{c}{FEDformer} \\
 \multicolumn{2}{c|}{} & \multicolumn{2}{c}{(ours)} & \multicolumn{2}{c}{(2025)} & \multicolumn{2}{c}{(2025)} & \multicolumn{2}{c}{(2025)} & \multicolumn{2}{c}{(2024)} & \multicolumn{2}{c}{(2024)} & \multicolumn{2}{c}{(2023)} & \multicolumn{2}{c}{(2023)} & \multicolumn{2}{c}{(2023)} & \multicolumn{2}{c}{(2023)} & \multicolumn{2}{c}{(2022)} & \multicolumn{2}{c}{(2022)} \\
\addlinespace\cline{1-26} \addlinespace
\multicolumn{2}{c|}{Metrics} & mse & mae & mse & mae & mse & mae & mse & mae & mse & mae & mse & mae & mse & mae & mse & mae & mse & mae & mse & mae & mse & mae & mse & mae \\
\midrule
        \multirow[c]{5}{*}{\rotatebox{90}{ETTh1}} & 96 & \textcolor{red}{\textbf{0.361}} & \textcolor{red}{\textbf{0.394}} & 0.370 & \textcolor{blue}{\underline{0.396}} & 0.373 & 0.399 & \textcolor{blue}{\underline{0.367}} & 0.397 & 0.386 & 0.405 & 0.372 & 0.401 & 0.377 & 0.397 & 0.411 & 0.435 & 0.389 & 0.412 & 0.379 & 0.403 & 0.591 & 0.524 & 0.379 & 0.419 \\
        ~ & 192 & \textcolor{red}{\textbf{0.401}} & \textcolor{blue}{\underline{0.418}} & \textcolor{blue}{\underline{0.403}} & \textcolor{red}{\textbf{0.417}} & 0.414 & 0.420 & 0.411 & 0.427 & 0.424 & 0.440 & 0.413 & 0.430 & 0.409 & 0.425 & 0.409 & 0.438 & 0.440 & 0.443 & 0.408 & 0.419 & 0.615 & 0.540 & 0.420 & 0.444 \\
        ~ & 336 & \textcolor{blue}{\underline{0.423}} & \textcolor{blue}{\underline{0.436}} & \textcolor{red}{\textbf{0.420}} & \textcolor{red}{\textbf{0.432}} & 0.442 & 0.446 & 0.436 & 0.442 & 0.449 & 0.460 & 0.438 & 0.450 & 0.431 & 0.444 & 0.433 & 0.457 & 0.523 & 0.487 & 0.440 & 0.440 & 0.632 & 0.551 & 0.458 & 0.466 \\
        ~ & 720 & \textcolor{red}{\textbf{0.433}} & \textcolor{red}{\textbf{0.453}} & \textcolor{blue}{\underline{0.442}} & \textcolor{blue}{\underline{0.463}} & 0.455 & 0.467 & 0.467 & 0.478 & 0.495 & 0.487 & 0.486 & 0.484 & 0.457 & 0.477 & 0.501 & 0.514 & 0.521 & 0.495 & 0.471 & 0.493 & 0.828 & 0.658 & 0.474 & 0.488 \\
        \cmidrule(lr){2-26}
        ~ & \textit{Avg} & \textcolor{red}{\textbf{0.405}} & \textcolor{red}{\textbf{0.425}} & \textcolor{blue}{\underline{0.409}} & \textcolor{blue}{\underline{0.427}} & 0.421 & 0.433 & 0.420 & 0.436 & 0.439 & 0.448 & 0.427 & 0.441 & 0.419 & 0.436 & 0.439 & 0.461 & 0.468 & 0.459 & 0.424 & 0.439 & 0.666 & 0.568 & 0.433 & 0.454 \\
        \addlinespace\cline{1-26} \addlinespace
        \multirow[c]{5}{*}{\rotatebox{90}{ETTh2}} & 96 & \textcolor{blue}{\underline{0.275}} & \textcolor{red}{\textbf{0.335}} & 0.280 & 0.343 & 0.287 & 0.349 & 0.276 & 0.344 & 0.297 & 0.348 & 0.281 & 0.351 & \textcolor{red}{\textbf{0.274}} & \textcolor{blue}{\underline{0.337}} & 0.728 & 0.603 & 0.334 & 0.370 & 0.300 & 0.364 & 0.347 & 0.387 & 0.337 & 0.380 \\
        ~ & 192 & 0.351 & \textcolor{red}{\textbf{0.381}} & \textcolor{red}{\textbf{0.329}} & \textcolor{blue}{\underline{0.382}} & 0.348 & 0.393 & \textcolor{blue}{\underline{0.347}} & 0.393 & 0.372 & 0.403 & 0.349 & 0.387 & 0.348 & 0.384 & 0.723 & 0.607 & 0.404 & 0.413 & 0.387 & 0.423 & 0.379 & 0.418 & 0.415 & 0.428 \\
        ~ & 336 & 0.382 & \textcolor{red}{\textbf{0.409}} & 0.370 & \textcolor{blue}{\underline{0.412}} & 0.383 & 0.423 & 0.376 & 0.425 & 0.388 & 0.417 & \textcolor{blue}{\underline{0.366}} & 0.413 & 0.377 & 0.416 & 0.740 & 0.628 & 0.389 & 0.435 & 0.490 & 0.487 & \textcolor{red}{\textbf{0.358}} & 0.413 & 0.389 & 0.457 \\
        ~ & 720 & \textcolor{red}{\textbf{0.399}} & \textcolor{red}{\textbf{0.432}} & 0.420 & 0.450 & 0.407 & 0.444 & 0.436 & 0.473 & 0.424 & 0.444 & \textcolor{blue}{\underline{0.401}} & \textcolor{blue}{\underline{0.436}} & 0.406 & 0.441 & 1.386 & 0.882 & 0.434 & 0.448 & 0.704 & 0.597 & 0.422 & 0.457 & 0.483 & 0.488 \\
        \cmidrule(lr){2-26}
        ~ & \textit{Avg} & 0.352 & \textcolor{red}{\textbf{0.389}} & \textcolor{blue}{\underline{0.350}} & 0.397 & 0.356 & 0.402 & 0.359 & 0.409 & 0.370 & 0.403 & \textcolor{red}{\textbf{0.349}} & 0.397 & 0.351 & \textcolor{blue}{\underline{0.395}} & 0.894 & 0.680 & 0.390 & 0.416 & 0.470 & 0.468 & 0.377 & 0.419 & 0.406 & 0.438 \\
        \addlinespace\cline{1-26} \addlinespace
        \multirow[c]{5}{*}{\rotatebox{90}{ETTm1}} & 96 & \textcolor{red}{\textbf{0.286}} & \textcolor{red}{\textbf{0.332}} & 0.290 & 0.348 & 0.292 & 0.346 & 0.302 & 0.349 & 0.300 & 0.353 & 0.293 & 0.345 & \textcolor{blue}{\underline{0.289}} & \textcolor{blue}{\underline{0.343}} & 0.314 & 0.367 & 0.340 & 0.378 & 0.300 & 0.345 & 0.415 & 0.410 & 0.463 & 0.463 \\
        ~ & 192 & \textcolor{red}{\textbf{0.325}} & \textcolor{red}{\textbf{0.359}} & 0.332 & 0.368 & \textcolor{blue}{\underline{0.327}} & \textcolor{blue}{\underline{0.365}} & 0.329 & 0.367 & 0.341 & 0.380 & 0.335 & 0.372 & 0.329 & 0.368 & 0.374 & 0.410 & 0.392 & 0.404 & 0.336 & 0.366 & 0.494 & 0.451 & 0.575 & 0.516 \\
        ~ & 336 & 0.359 & \textcolor{red}{\textbf{0.379}} & \textcolor{red}{\textbf{0.354}} & 0.386 & 0.365 & 0.386 & \textcolor{blue}{\underline{0.355}} & \textcolor{blue}{\underline{0.383}} & 0.374 & 0.396 & 0.368 & 0.386 & 0.362 & 0.390 & 0.413 & 0.432 & 0.423 & 0.426 & 0.367 & 0.386 & 0.577 & 0.490 & 0.618 & 0.544 \\
        ~ & 720 & 0.417 & \textcolor{red}{\textbf{0.410}} & \textcolor{red}{\textbf{0.401}} & 0.417 & 0.427 & 0.419 & \textcolor{blue}{\underline{0.406}} & \textcolor{blue}{\underline{0.413}} & 0.429 & 0.430 & 0.426 & 0.417 & 0.416 & 0.423 & 0.753 & 0.613 & 0.475 & 0.453 & 0.419 & 0.416 & 0.636 & 0.535 & 0.612 & 0.551 \\
        \cmidrule(lr){2-26}
        ~ & \textit{Avg} & \textcolor{blue}{\underline{0.347}} & \textcolor{red}{\textbf{0.370}} & \textcolor{red}{\textbf{0.344}} & 0.380 & 0.353 & 0.379 & 0.348 & \textcolor{blue}{\underline{0.378}} & 0.361 & 0.390 & 0.355 & 0.380 & 0.349 & 0.381 & 0.464 & 0.455 & 0.407 & 0.415 & 0.356 & \textcolor{blue}{\underline{0.378}} & 0.530 & 0.472 & 0.567 & 0.519 \\
        \addlinespace\cline{1-26} \addlinespace
        \multirow[c]{5}{*}{\rotatebox{90}{ETTm2}} & 96 & \textcolor{red}{\textbf{0.164}} & \textcolor{red}{\textbf{0.248}} & \textcolor{red}{\textbf{0.164}} & \textcolor{blue}{\underline{0.254}} & \textcolor{red}{\textbf{0.164}} & \textcolor{blue}{\underline{0.254}} & \textcolor{red}{\textbf{0.164}} & 0.256 & 0.175 & 0.266 & \textcolor{blue}{\underline{0.165}} & 0.256 & \textcolor{blue}{\underline{0.165}} & 0.255 & 0.296 & 0.391 & 0.189 & 0.265 & \textcolor{red}{\textbf{0.164}} & 0.255 & 0.210 & 0.294 & 0.216 & 0.309 \\
        ~ & 192 & \textcolor{red}{\textbf{0.219}} & \textcolor{red}{\textbf{0.286}} & 0.238 & 0.300 & 0.226 & 0.300 & \textcolor{red}{\textbf{0.219}} & 0.296 & 0.242 & 0.312 & 0.225 & 0.298 & \textcolor{blue}{\underline{0.221}} & \textcolor{blue}{\underline{0.293}} & 0.369 & 0.416 & 0.254 & 0.310 & 0.224 & 0.304 & 0.338 & 0.373 & 0.297 & 0.360 \\
        ~ & 336 & \textcolor{red}{\textbf{0.268}} & \textcolor{red}{\textbf{0.321}} & 0.278 & 0.331 & 0.276 & 0.331 & \textcolor{blue}{\underline{0.275}} & 0.336 & 0.282 & 0.337 & 0.277 & 0.332 & 0.276 & \textcolor{blue}{\underline{0.327}} & 0.588 & 0.600 & 0.313 & 0.345 & 0.277 & 0.337 & 0.432 & 0.416 & 0.366 & 0.400 \\
        ~ & 720 & \textcolor{red}{\textbf{0.346}} & \textcolor{red}{\textbf{0.372}} & 0.359 & 0.387 & \textcolor{blue}{\underline{0.358}} & 0.388 & 0.359 & 0.392 & 0.375 & 0.394 & 0.360 & 0.387 & 0.362 & \textcolor{blue}{\underline{0.381}} & 0.750 & 0.612 & 0.413 & 0.402 & 0.371 & 0.401 & 0.554 & 0.476 & 0.459 & 0.450 \\
        \cmidrule(lr){2-26}
        ~ & \textit{Avg} & \textcolor{red}{\textbf{0.249}} & \textcolor{red}{\textbf{0.307}} & 0.260 & 0.318 & 0.256 & 0.318 & \textcolor{blue}{\underline{0.254}} & 0.320 & 0.268 & 0.327 & 0.257 & 0.318 & 0.256 & \textcolor{blue}{\underline{0.314}} & 0.501 & 0.505 & 0.292 & 0.331 & 0.259 & 0.324 & 0.384 & 0.390 & 0.335 & 0.380 \\
        \addlinespace\cline{1-26} \addlinespace
        \multirow[c]{5}{*}{\rotatebox{90}{Weather}} & 96 & \textcolor{blue}{\underline{0.146}} & \textcolor{red}{\textbf{0.189}} & 0.151 & 0.202 & 0.147 & 0.199 & 0.165 & 0.222 & 0.157 & 0.207 & 0.147 & 0.198 & 0.149 & \textcolor{blue}{\underline{0.196}} & \textcolor{red}{\textbf{0.143}} & 0.210 & 0.168 & 0.214 & 0.170 & 0.230 & 0.188 & 0.242 & 0.229 & 0.298 \\
        ~ & 192 & \textcolor{red}{\textbf{0.190}} & \textcolor{red}{\textbf{0.230}} & 0.195 & 0.244 & 0.194 & 0.245 & 0.211 & 0.264 & 0.200 & 0.248 & 0.192 & 0.243 & \textcolor{blue}{\underline{0.191}} & \textcolor{blue}{\underline{0.239}} & 0.198 & 0.260 & 0.219 & 0.262 & 0.216 & 0.273 & 0.241 & 0.290 & 0.265 & 0.334 \\
        ~ & 336 & \textcolor{red}{\textbf{0.241}} & \textcolor{red}{\textbf{0.271}} & \textcolor{blue}{\underline{0.242}} & 0.287 & 0.243 & 0.282 & 0.260 & 0.302 & 0.252 & 0.287 & 0.247 & 0.284 & \textcolor{blue}{\underline{0.242}} & \textcolor{blue}{\underline{0.279}} & 0.258 & 0.314 & 0.278 & 0.302 & 0.258 & 0.307 & 0.341 & 0.341 & 0.330 & 0.372 \\
        ~ & 720 & 0.315 & \textcolor{red}{\textbf{0.322}} & 0.317 & 0.340 & \textcolor{red}{\textbf{0.310}} & \textcolor{blue}{\underline{0.329}} & 0.327 & 0.355 & 0.320 & 0.336 & 0.318 & 0.330 & \textcolor{blue}{\underline{0.312}} & 0.330 & 0.335 & 0.385 & 0.353 & 0.351 & 0.323 & 0.362 & 0.403 & 0.388 & 0.423 & 0.418 \\
        \cmidrule(lr){2-26}
        ~ & \textit{Avg} & \textcolor{red}{\textbf{0.223}} & \textcolor{red}{\textbf{0.253}} & \textcolor{blue}{\underline{0.226}} & 0.268 & \textcolor{red}{\textbf{0.223}} & 0.264 & 0.241 & 0.286 & 0.232 & 0.270 & \textcolor{blue}{\underline{0.226}} & 0.264 & \textcolor{red}{\textbf{0.223}} & \textcolor{blue}{\underline{0.261}} & 0.233 & 0.292 & 0.255 & 0.282 & 0.242 & 0.293 & 0.293 & 0.315 & 0.312 & 0.355 \\
        \addlinespace\cline{1-26} \addlinespace
        \multirow[c]{5}{*}{\rotatebox{90}{Electricity}} & 96 & \textcolor{red}{\textbf{0.127}} & \textcolor{red}{\textbf{0.218}} & 0.135 & 0.231 & \textcolor{blue}{\underline{0.132}} & \textcolor{blue}{\underline{0.227}} & 0.133 & 0.232 & 0.134 & 0.230 & 0.153 & 0.256 & 0.143 & 0.247 & 0.134 & 0.231 & 0.169 & 0.271 & 0.140 & 0.237 & 0.171 & 0.274 & 0.191 & 0.305 \\
        ~ & 192 & \textcolor{blue}{\underline{0.147}} & \textcolor{red}{\textbf{0.236}} & 0.149 & \textcolor{blue}{\underline{0.243}} & 0.149 & \textcolor{blue}{\underline{0.243}} & 0.149 & 0.247 & 0.154 & 0.250 & 0.168 & 0.269 & 0.158 & 0.260 & \textcolor{red}{\textbf{0.146}} & \textcolor{blue}{\underline{0.243}} & 0.180 & 0.280 & 0.154 & 0.251 & 0.180 & 0.283 & 0.203 & 0.316 \\
        ~ & 336 & \textcolor{blue}{\underline{0.162}} & \textcolor{red}{\textbf{0.254}} & 0.165 & 0.260 & 0.167 & 0.261 & \textcolor{red}{\textbf{0.161}} & \textcolor{blue}{\underline{0.259}} & 0.169 & 0.265 & 0.189 & 0.291 & 0.168 & 0.267 & 0.165 & 0.264 & 0.204 & 0.304 & 0.169 & 0.268 & 0.204 & 0.305 & 0.221 & 0.333 \\
        ~ & 720 & 0.199 & \textcolor{red}{\textbf{0.286}} & 0.206 & 0.297 & 0.203 & 0.292 & \textcolor{blue}{\underline{0.197}} & 0.297 & \textcolor{red}{\textbf{0.194}} & \textcolor{blue}{\underline{0.288}} & 0.228 & 0.320 & 0.214 & 0.307 & 0.237 & 0.314 & 0.205 & 0.304 & 0.204 & 0.301 & 0.221 & 0.319 & 0.259 & 0.364 \\
        \cmidrule(lr){2-26}
        ~ & \textit{Avg} & \textcolor{red}{\textbf{0.159}} & \textcolor{red}{\textbf{0.248}} & 0.164 & 0.258 & 0.163 & \textcolor{blue}{\underline{0.256}} & \textcolor{blue}{\underline{0.160}} & 0.259 & 0.163 & 0.258 & 0.184 & 0.284 & 0.171 & 0.270 & 0.171 & 0.263 & 0.189 & 0.290 & 0.167 & 0.264 & 0.194 & 0.295 & 0.218 & 0.330 \\
        \addlinespace\cline{1-26} \addlinespace
        \multirow[c]{5}{*}{\rotatebox{90}{Solar}} & 96 & \textcolor{red}{\textbf{0.169}} & \textcolor{red}{\textbf{0.202}} & 0.187 & 0.255 & 0.175 & 0.237 & 0.192 & 0.251 & 0.190 & 0.244 & 0.179 & 0.232 & \textcolor{blue}{\underline{0.170}} & 0.234 & 0.183 & \textcolor{blue}{\underline{0.208}} & 0.198 & 0.270 & 0.199 & 0.265 & 0.381 & 0.398 & 0.485 & 0.570 \\
        ~ & 192 & \textcolor{red}{\textbf{0.189}} & \textcolor{red}{\textbf{0.214}} & 0.194 & 0.265 & 0.198 & 0.259 & 0.247 & 0.323 & \textcolor{blue}{\underline{0.193}} & 0.257 & 0.201 & 0.259 & 0.204 & 0.302 & 0.208 & \textcolor{blue}{\underline{0.226}} & 0.206 & 0.276 & 0.220 & 0.282 & 0.395 & 0.386 & 0.415 & 0.477 \\
        ~ & 336 & \textcolor{blue}{\underline{0.195}} & \textcolor{red}{\textbf{0.220}} & 0.203 & 0.264 & 0.213 & 0.259 & 0.240 & 0.300 & 0.203 & 0.266 & \textcolor{red}{\textbf{0.190}} & 0.256 & 0.212 & 0.293 & 0.212 & \textcolor{blue}{\underline{0.239}} & 0.208 & 0.284 & 0.234 & 0.295 & 0.410 & 0.394 & 1.008 & 0.839 \\
        ~ & 720 & \textcolor{red}{\textbf{0.199}} & \textcolor{red}{\textbf{0.225}} & 0.209 & 0.269 & 0.222 & 0.269 & 0.246 & 0.311 & 0.223 & 0.281 & \textcolor{blue}{\underline{0.203}} & 0.261 & 0.215 & 0.307 & 0.215 & \textcolor{blue}{\underline{0.256}} & 0.232 & 0.294 & 0.243 & 0.301 & 0.377 & 0.376 & 0.655 & 0.627 \\
        \cmidrule(lr){2-26}
        ~ & \textit{Avg} & \textcolor{red}{\textbf{0.188}} & \textcolor{red}{\textbf{0.215}} & 0.198 & 0.263 & 0.202 & 0.256 & 0.231 & 0.296 & 0.202 & 0.262 & \textcolor{blue}{\underline{0.193}} & 0.252 & 0.200 & 0.284 & 0.204 & \textcolor{blue}{\underline{0.232}} & 0.211 & 0.281 & 0.224 & 0.286 & 0.391 & 0.388 & 0.641 & 0.628 \\
        \addlinespace\cline{1-26} \addlinespace
        \multirow[c]{5}{*}{\rotatebox{90}{Traffic}} & 96 & \textcolor{red}{\textbf{0.363}} & \textcolor{red}{\textbf{0.241}} & 0.388 & 0.269 & 0.391 & 0.277 & 0.378 & 0.273 & \textcolor{red}{\textbf{0.363}} & 0.265 & \textcolor{blue}{\underline{0.369}} & \textcolor{blue}{\underline{0.257}} & 0.370 & 0.262 & 0.526 & 0.288 & 0.595 & 0.312 & 0.395 & 0.275 & 0.604 & 0.330 & 0.593 & 0.365 \\
        ~ & 192 & \textcolor{red}{\textbf{0.383}} & \textcolor{red}{\textbf{0.251}} & 0.411 & 0.286 & 0.405 & 0.283 & 0.391 & 0.277 & \textcolor{blue}{\underline{0.384}} & 0.273 & 0.400 & 0.272 & 0.386 & 0.269 & 0.503 & \textcolor{blue}{\underline{0.263}} & 0.613 & 0.322 & 0.407 & 0.280 & 0.610 & 0.338 & 0.614 & 0.381 \\
        ~ & 336 & \textcolor{red}{\textbf{0.394}} & \textcolor{red}{\textbf{0.257}} & 0.425 & 0.284 & 0.416 & 0.290 & 0.402 & 0.282 & \textcolor{blue}{\underline{0.396}} & 0.277 & 0.407 & \textcolor{blue}{\underline{0.272}} & \textcolor{blue}{\underline{0.396}} & 0.275 & 0.505 & 0.276 & 0.626 & 0.332 & 0.417 & 0.286 & 0.626 & 0.341 & 0.627 & 0.389 \\
        ~ & 720 & \textcolor{red}{\textbf{0.431}} & \textcolor{red}{\textbf{0.278}} & 0.455 & 0.302 & 0.454 & 0.312 & \textcolor{blue}{\underline{0.434}} & 0.297 & 0.445 & 0.308 & 0.461 & 0.316 & 0.435 & \textcolor{blue}{\underline{0.295}} & 0.552 & 0.301 & 0.635 & 0.340 & 0.454 & 0.308 & 0.643 & 0.347 & 0.646 & 0.394 \\
        \cmidrule(lr){2-26}
        ~ & \textit{Avg} & \textcolor{red}{\textbf{0.393}} & \textcolor{red}{\textbf{0.257}} & 0.420 & 0.285 & 0.416 & 0.291 & 0.401 & 0.282 & \textcolor{blue}{\underline{0.397}} & 0.281 & 0.409 & 0.279 & \textcolor{blue}{\underline{0.397}} & \textcolor{blue}{\underline{0.275}} & 0.521 & 0.282 & 0.617 & 0.327 & 0.418 & 0.287 & 0.621 & 0.339 & 0.620 & 0.382 \\
        \addlinespace
        \bottomrule
    \end{tabular}
}
\end{table*}
\subsection{Ablation Studies}
To examine the contribution of each hierarchy loss, we conduct ablation studies on four datasets that span both small-scale (ETTh1, ETTh2) and large-scale (Solar, Traffic) settings. For each dataset, we report MSE at two representative prediction horizons, $H{=}96$ and $H{=}720$, corresponding to short-term and long-term forecasting respectively. This setup allows us to assess whether the proposed losses generalize across datasets of varying scale and across different forecasting ranges. Results are summarized in Table~\ref{tab:ablation}.

\paragraph{Loss complementarity.}
The radial ordering loss $\mathcal{L}_r$ generally provides the larger individual contribution, particularly on ETTh2, where $\mathcal{L}_r$ alone nearly recovers the full model's improvement (e.g., 0.410$\to$0.399 at $H{=}720$). This suggests that radial depth ordering is a key mechanism for encoding the coarse-to-fine hierarchy. However, the relative contribution varies across datasets, with $\mathcal{L}_a$ providing greater standalone improvements in some settings such as Traffic.

More importantly, the two losses are most effective when combined. On ETTh1 at $H{=}720$, for example, neither loss alone yields a meaningful improvement, whereas their combination reduces MSE from 0.443 to 0.433. Similar patterns are observed on Solar. These results demonstrate the complementarity of the two constraints: $\mathcal{L}_r$ organizes representations along the radial axis, while $\mathcal{L}_a$ structures their angular relationships, jointly forming a hierarchical representation that neither constraint consistently achieves alone.

\paragraph{Effect grows with horizon and dataset scale.}
The benefit of hierarchy losses is generally larger at the long-term horizon ($H{=}720$) than at the short-term horizon ($H{=}96$), and is most pronounced on the large-scale Solar dataset, where the full model reduces MSE by 9.1\% at $H{=}720$. Even on Traffic, a 862-channel dataset where channel-wise variance often overshadows temporal structure, hierarchy losses still yield consistent improvements, reaching 1.4\% MSE reduction at $H{=}720$ and reinforcing the pattern that longer horizons benefit more from explicit hierarchical structure.

\begin{table}[t]
\centering
\caption{Ablation on small-scale (ETTh1, ETTh2) and large-scale (Solar, Traffic) datasets at short-term ($H{=}96$) and long-term ($H{=}720$) horizons. All values are MSE.}
\label{tab:ablation}
\renewcommand{\arraystretch}{1.1}
\begin{tabular}{cl|c|c|c|c}
\toprule
Data & $H$ & None & $\mathcal{L}_r$ only & $\mathcal{L}_a$ only & Full Model \\
\midrule
\multirow{2}{*}{ETTh1}
  & 96  & 0.374 & 0.364 & 0.370 & \textbf{0.361} \\
  & 720 & 0.443 & 0.442 & 0.443 & \textbf{0.433} \\
\midrule
\multirow{2}{*}{ETTh2}
  & 96  & 0.279 & \textbf{0.275} & 0.276 & \textbf{0.275} \\
  & 720 & 0.410 & \textbf{0.399} & 0.408 & \textbf{0.399} \\
\midrule
\multirow{2}{*}{Solar}
  & 96  & 0.176 & 0.170 & 0.171 & \textbf{0.169} \\
  & 720 & 0.219 & 0.200 & 0.201 & \textbf{0.199} \\
\midrule
\multirow{2}{*}{Traffic}
  & 96  & 0.364 & 0.364 & 0.363 & \textbf{0.363} \\
  & 720 & 0.437 & 0.432 & 0.434 & \textbf{0.431} \\
\bottomrule
\end{tabular}%
\end{table}

\subsection{Emergence of Hierarchical Structure.}
\begin{figure}[htbp]
    \centering
    \subfloat[ETTh1 depth emergence]{\includegraphics[width=0.45\linewidth]{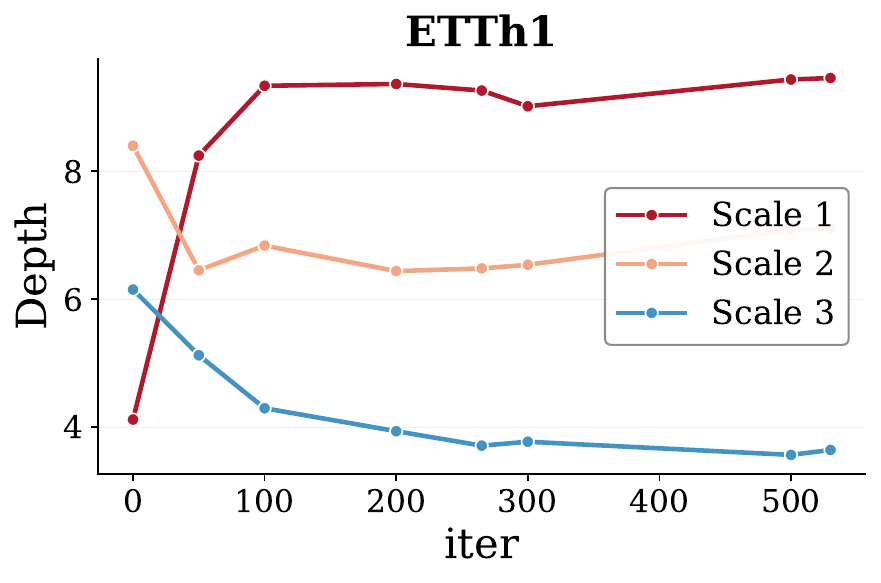}}\hspace{0.06\linewidth}
    \subfloat[ETTh2 depth emergence]{\includegraphics[width=0.45\linewidth]{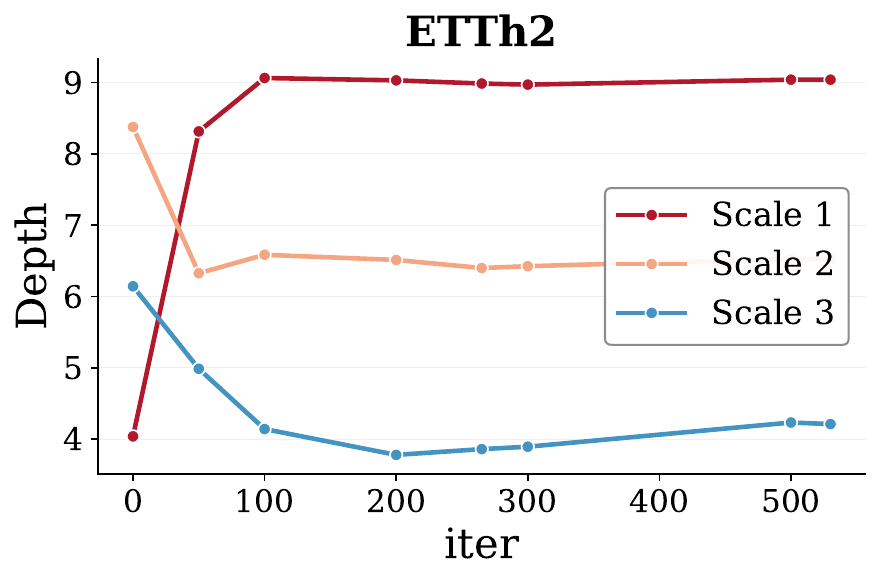}}\\[4pt]
    \subfloat[ETTh1 radial depth]{\includegraphics[width=0.45\linewidth]{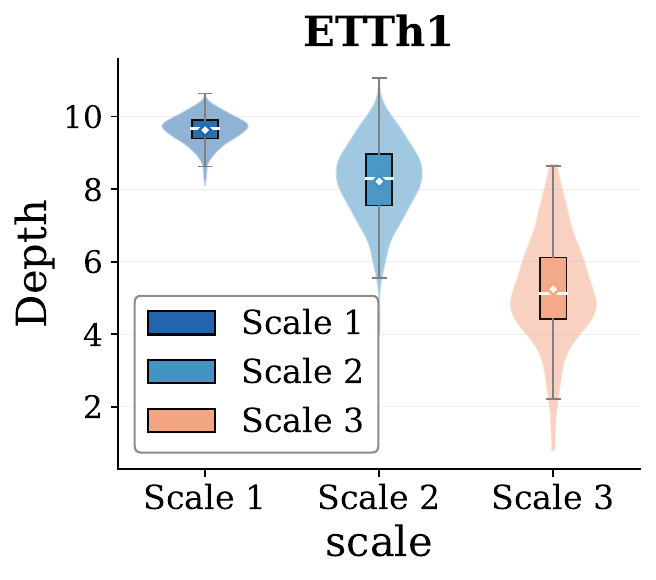}}\hspace{0.06\linewidth}
    \subfloat[ETTh2 radial depth]{\includegraphics[width=0.45\linewidth]{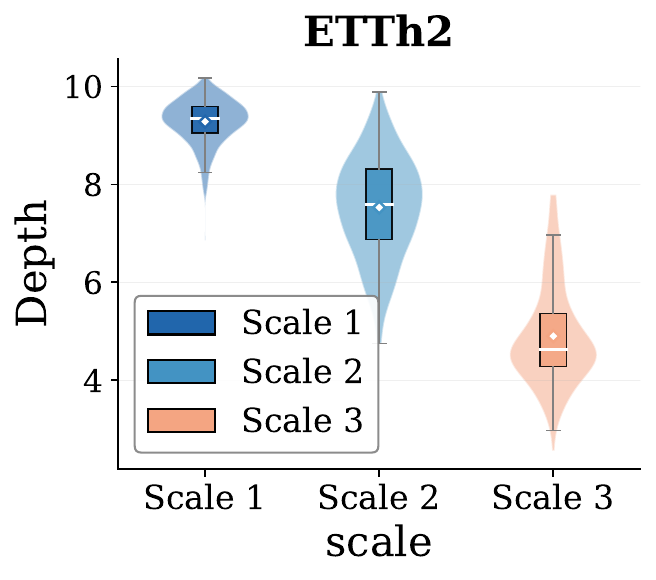}}
    \caption{Emergence of hierarchical structure. (a)--(b) Per-scale mean hyperbolic depth over training iterations on ETTh1 and ETTh2. (c)--(d) Distribution of hyperbolic depths by scale on the test set after training.}
    \label{fig:hierarchy_emergence}
\end{figure}

To examine how the hierarchy losses shape the embedding geometry during training, we visualize the per-scale hyperbolic depth across training iterations. Figure~\ref{fig:hierarchy_emergence}(a)--(b) plot the mean depth $d_c(\mathbf{0}, \mathbf{z}^{(s)})$ for each scale $s$, measured on the test set at regular intervals on ETTh1 and ETTh2 over 2 training epochs.

At initialization, there is no hierarchical separation. As training progresses, a clear depth ordering emerges. The final depth ordering, Scale 1 $>$ Scale 2 $>$ Scale 3, confirms that the hierarchy losses successfully induce the intended structure. Finer temporal resolutions are embedded deeper in the Poincar\'e ball, closer to the boundary, whereas coarser resolutions remain shallow near the origin. Figure~\ref{fig:hierarchy_emergence}(c)--(d) further illustrate this separation by showing the per-scale distribution of $d_c(\mathbf{0}, \mathbf{z})$ on the test set after training, where the scales occupy clearly distinct radial bands with minimal overlap.

\subsection{Advantage of Hyperbolic Space}

To further investigate how effective our choice of modeling the multi-scale hierarchy in hyperbolic space is, we conduct a controlled comparison against an Euclidean counterpart. This experiment directly tests whether the geometric inductive bias of hyperbolic space contributes to forecasting performance beyond what an architecturally identical Euclidean variant can achieve.

\paragraph{Controlled comparison.}
To isolate the effect of geometry from model capacity, we construct a Euclidean baseline by replacing HypLinear (Eq. \ref{eq:hyplinear}) with a standard Linear layer, while keeping all other architectural components unchanged. Since the two layers share identical weight and bias dimensions, the total number of learnable parameters remains the same. For the hierarchy-preserving losses, we substitute their Euclidean counterparts, using norm-based depth and cosine-based angular alignment. This controlled setting isolates the effect of the representation geometry from differences in model capacity.

\paragraph{Hyperbolic consistently outperforms Euclidean.}
As shown in Table~\ref{tab:hyp_vs_euc}, the hyperbolic model consistently outperforms its Euclidean counterpart, with larger differences at longer prediction horizons. On Solar, the MSE improvement increases from 1.0\% at $H{=}192$ to 9.5\% at $H{=}720$ (0.220$\to$0.199), while on ETTh1 it increases from 1.0\% to 4.0\% (0.451$\to$0.433). Since the two variants differ only in the representation geometry, these results support the use of hyperbolic space for modeling the proposed cross-scale hierarchy, particularly for longer-horizon forecasting.

\begin{table}[t]
\centering
\caption{Comparison between hyperbolic and Euclidean representations. All values are MSE.}
\label{tab:hyp_vs_euc}

\resizebox{0.95\columnwidth}{!}{
\begin{tabular}{c|cc|cc|cc}
\toprule

& \multicolumn{2}{c|}{Solar}
& \multicolumn{2}{c|}{ETTh1}
& \multicolumn{2}{c}{Weather} \\

\cmidrule(lr){2-3}
\cmidrule(lr){4-5}
\cmidrule(lr){6-7}

$H$
& Hyperbolic & Euclidean
& Hyperbolic & Euclidean
& Hyperbolic & Euclidean \\

\midrule

192
& \textbf{0.189} & 0.191
& \textbf{0.401} & 0.405
& \textbf{0.190} & 0.196 \\

336
& \textbf{0.195} & 0.199
& \textbf{0.423} & 0.424
& \textbf{0.241} & 0.243 \\

720
& \textbf{0.199} & 0.220
& \textbf{0.433} & 0.451
& \textbf{0.315} & 0.318 \\

\bottomrule
\end{tabular}
}
\end{table}

\subsection{Plug-and-Play Experiment}
In this analysis, we integrate the core components of HypLTSF---the hyperbolic embedding layer and the temporal hierarchy losses ($\mathcal{L}_r$, $\mathcal{L}_a$)---into two representative yet architecturally distinct multi-scale forecasting models: TimeMixer~\cite{timemixer}, which constructs its multi-scale representation through progressive downsampling in the time domain, and MICN~\cite{micn}, which derives multi-scale features via learned convolution maps at different kernel sizes. The two models cover complementary design philosophies, namely temporal pooling and convolutional filtering, making them a suitable test bed for evaluating the generalizability of our approach across different multi-scale paradigms. 

As shown in Table~\ref{tab:plugin_result}, applying the proposed module improves performance in most settings. On ETTm1 at $H=96$, for example, MSE decreases from 0.293 to 0.288 for TimeMixer and from 0.303 to 0.289 for MICN. A few exceptions occur at longer horizons, such as TimeMixer at $H=336$, where MSE slightly increases while MAE decreases.

Overall, the results demonstrate that the hyperbolic embedding and hierarchy losses can serve as an architecture-agnostic plug-in that transfers to different multi-scale strategies. Performance improvements are observed in most cases, although the magnitude varies across base models and prediction horizons.

\begin{table}[!t]
\caption{Full results of the plug-and-play long-term forecasting experiments with HypLTSF, evaluated at prediction horizons $H \in \{96, 192, 336, 720\}$.}
\label{tab:plugin_result}
\setlength{\tabcolsep}{4pt}
\scriptsize
\centering
\begin{threeparttable}

\begin{tabular}{cc|cc|cc|cc|cc}

\toprule
 \multicolumn{2}{c|}{\scalebox{1.1}{Models}} & \multicolumn{2}{c|}{TimeMixer} & \multicolumn{2}{c|}{+HypLTSF} & \multicolumn{2}{c|}{MICN} & \multicolumn{2}{c}{+HypLTSF} \\

 \cmidrule(lr){3-4} \cmidrule(lr){5-6} \cmidrule(lr){7-8} \cmidrule(lr){9-10}

 \multicolumn{1}{c}{Dataset} & \multicolumn{1}{c|}{H} & \scalebox{0.9}{MSE} & \scalebox{0.9}{MAE} & \scalebox{0.9}{MSE} & \scalebox{0.9}{MAE} & \scalebox{0.9}{MSE} & \scalebox{0.9}{MAE} & \scalebox{0.9}{MSE} & \scalebox{0.9}{MAE} \\

\toprule

\multirow{4}{*}{\rotatebox{90}{ETTm1}} & 96  & 0.293 & 0.345  & 0.288 & 0.332 & 0.303 & 0.349 & 0.289 & 0.338 \\
                       & 192 & 0.335 & 0.372 & 0.332 & 0.358 & 0.336 & 0.369 & 0.333 & 0.363 \\
                       & 336 & 0.368 & 0.386 & 0.370 & 0.379 & 0.370 & 0.391 & 0.369 & 0.390 \\
                       & 720 & 0.426 & 0.417 & 0.424 & 0.411 & 0.410 & 0.421 & 0.412 & 0.411 \\

\midrule

\multirow{4}{*}{\rotatebox{90}{Weather}} & 96  & 0.147 & 0.198 & 0.148 & 0.189 & 0.172 & 0.232 & 0.164 & 0.210 \\
                         & 192 & 0.191 & 0.242 & 0.190 & 0.231 & 0.214 & 0.271 & 0.212 & 0.253 \\
                         & 336 & 0.244 & 0.280 & 0.242 & 0.280 & 0.259 & 0.309 & 0.256 & 0.286 \\
                         & 720 & 0.316 & 0.331 & 0.315 & 0.330 & 0.309 & 0.343 & 0.310 & 0.345 \\
\bottomrule
\end{tabular}


\vspace{-2em}

\end{threeparttable}
\end{table}

\subsection{Model Efficiency Analysis}
\begin{table}[htbp]
\centering
\caption{Efficiency and performance comparison on Traffic and Solar datasets. Best results are in \textbf{bold}.}
\label{tab:efficiency}
\resizebox{0.95\columnwidth}{!}{
\begin{tabular}{cl|ccccc}
\toprule
Dataset & Model & Memory (MB) $\downarrow$ & Infer (ms) $\downarrow$ & Train (ms) $\downarrow$ & MSE $\downarrow$ & MAE $\downarrow$ \\
\midrule
\multirow{3}{*}{\rotatebox{90}{Traffic}}
& TimeKAN   & \textbf{2870.6} & 414.42          & 632.85          & 0.411          & 0.286          \\
& TimeMixer & 3070.2          & \textbf{134.78} & \textbf{337.34} & 0.400          & 0.271          \\
& HypLTSF & 3167.8          & 142.64          & 387.79          & \textbf{0.383} & \textbf{0.251} \\
\midrule
\multirow{3}{*}{\rotatebox{90}{Solar}}
& TimeKAN   & \textbf{469.2} & 37.65           & 95.29           & 0.203          & 0.264          \\
& TimeMixer & 507.8          & 22.93           & 59.44           & 0.214          & 0.272          \\
& HypLTSF & 497.2          & \textbf{20.67}  & \textbf{56.85}  & \textbf{0.195} & \textbf{0.219} \\
\bottomrule
\end{tabular}%
}
\end{table}
To evaluate the computational efficiency of HypLTSF, we compare memory consumption, training time, inference time, and forecasting performance against state-of-the-art multi-scale forecasting models on two large-scale datasets: Traffic with 862 variables and Solar with 137 variables. Table~\ref{tab:efficiency} presents the efficiency comparison across the two datasets. For Traffic, we use a look-back window of 512 and a forecasting horizon of 192, while for Solar, we use the same look-back window with a forecasting horizon of 336.

Overall, HypLTSF provides a favorable trade-off between forecasting accuracy and computational efficiency. On Traffic, it achieves the best forecasting accuracy with only moderate increases in training and inference time compared with TimeMixer. On Solar, HypLTSF achieves the best forecasting accuracy while also providing the fastest training and inference. Although TimeKAN maintains the lowest memory consumption, its computational latency is higher on both datasets. These results show that explicit hierarchical modeling improves forecasting accuracy while maintaining competitive computational efficiency.

\section{Conclusion}
In this work, we proposed HypLTSF, a geometry-aware framework that explicitly models the multi-scale hierarchy of time series by embedding scale-wise representations into the Poincar\'e ball. To shape the embedding geometry, we introduced two hierarchy-preserving losses: a radial ordering loss that enforces fine-scale embeddings to lie deeper than their coarse-scale counterparts, encoding parent--child relationships along the radial axis, and an angular coherence loss that encourages sibling time steps to cluster directionally, forming coherent branches in the embedding space. Together, these losses induce a tree-like structure in which depth encodes temporal resolution and angular proximity encodes local temporal context. We evaluate HypLTSF on eight real-world benchmark datasets. The results demonstrate state-of-the-art performance across diverse forecasting horizons and the learned embeddings exhibit the intended hierarchical structure, empirically validating the effectiveness of geometry-aware hierarchical modeling as an inductive bias.

\bibliographystyle{IEEEtran}
\bibliography{IEEEabrv,Bibliography}

\end{document}